\documentclass{article}
\usepackage{amsmath}
\usepackage{amssymb}
\usepackage{algorithm} 
\usepackage{algorithmic} 
\usepackage[algo2e]{algorithm2e} 
\usepackage{microtype}
\usepackage{graphicx}
\usepackage{subfigure}
\usepackage{booktabs} 
\usepackage{multirow}
\usepackage{siunitx}

\usepackage{makecell, caption, booktabs, multirow, siunitx}

\usepackage{hyperref}

\usepackage[accepted]{icml2020}

\begin{document}

\twocolumn[
\icmltitle{\textbf{ Asynchronously Trained Distributed Topographic Maps}}

\icmlsetsymbol{equal}{*}

\begin{icmlauthorlist}
\icmlauthor{Abbas Siddiqui}{add1}
\icmlauthor{Dionysios Georgiadis}{add2}

\end{icmlauthorlist}

\icmlaffiliation{add1}{Bavaria, Germany}
\icmlaffiliation{add2}{Aspect Capital, London, United Kingdom}

\icmlcorrespondingauthor{Abbas Siddiqui}{abbas.siddiqui@gmail.com}
\icmlcorrespondingauthor{Dionysios Georgiadis}{dionysios.georgiadis@aspectcapital.com}

\icmlkeywords{Machine Learning, Distributed Topographic Maps, Asynchronous Training}

\vskip 0.3in
]

\printAffiliationsAndNotice{}  

\begin{abstract}
Topographic feature maps are low dimensional representations of data, that preserve spatial dependencies. Current methods of training such maps (e.g. self organizing maps - SOM, generative topographic maps) require centralized control and synchronous execution, which restricts scalability. We present an algorithm that uses $N$ autonomous units to generate a feature map by distributed asynchronous training. Unit autonomy is achieved by sparse interaction in time \& space through the combination of a distributed heuristic search, and a cascade-driven weight updating scheme governed by two rules: a unit i) adapts when it receives either a sample, or the weight vector of a neighbor, and ii) broadcasts its weight vector to its neighbors after adapting for a predefined number of times. Thus, a vector update can trigger an avalanche of adaptation. We map avalanching to a statistical mechanics model, which allows us to parametrize the statistical properties of cascading. Using MNIST, we empirically investigate the effect of the heuristic search accuracy and the cascade parameters on map quality. We also provide empirical evidence that algorithm complexity scales at most linearly with system size $N$. The proposed approach is found to perform comparably with similar methods in classification tasks across multiple datasets.

\end{abstract}
\section{Introduction} \label{intro}

Topographic maps are low dimensional discrete representations of high dimensional data. These maps consist of neurons (a.k.a. units) interconnected over a predefined regular lattice.  
Each neuron $j$ contains a representation of the data $w_j$. 
When training such a map, the goal is twofold: 
i) the representation $w_j$ of each neuron should approximate an area of the data distribution, and 
ii) adjacent neurons should contain similar representations (respecting the the map's topology).  
Such maps have found broad application in data analytics, (mainly for clustering, function approximation, dimensionality reduction - see \cite{kohonen2013_essentials, kohonen_book} for more examples), and are typically trained via the Self Organizing Map algorithm (SOM). 

Scalability is increasingly needed in machine learning algorithms (due to growing data accessibility \cite{surveymachinelearning}).
However, the SOM requires centralized training - which compromises its scalability. 
The need for centralized training comes from the best matching unit (BMU) search (a centralized process in which a sample is compared with every unit's representation $w_j$), and the unit adaptation process (where a central entity has knowledge of all the units in the map). 
To tackle the scalability challenge many workarounds have been proposed, including  parallel execution, dedicated hardware acceleration (with GPUs \cite{parallelcudaSOM,gpumassively} or FPGAs \cite{FPGAsom}), and heuristic methods \cite{massivemap}. 
Crucially however, distributed implementations \cite{parallelSOM,parallelcudaSOM} of SOM rely on iterative map-reduce frameworks (Spark, HADOOP). 
Such frameworks require synchronous execution, and are thus susceptible to network latency, or slow workers (i.e. stranglers). Asynchronous training methods have attracted attention, because they do not suffer from such limitations \cite{mnih2016asynchronous,li2019generalized,fernando2017pathnet}.

In order to tackle the problem of scalability, in this work we present an algorithm for the training topographic maps through asynchronous execution. 
This is made possible by converting map units to autonomous agents. 
Each agent has limited knowledge of the feature map (a few neighbors), and interacts sparsely with its peers. 
The search and adjustment processes are asynchronous and require only the local knowledge of each unit's predefined neighbors.

Training involves two processes for each sample $s_i$. 
 First the a good-matching unit (GMU) is found through a heuristic search, and then the GMU adapts $w_j$ to the sample - potentially instigating an \textit{adaptation cascade} (as explained below). 
 For effective exploration, the search relies on non-local connections between units which are dubbed \textit{far links}. 
 Also, the search uses a set of local connections between units - dubbed \textit{near links} - for greedy exploitation. 
  The potential for adaptation cascades allows the map to preserve its topology during training. 
 Cascading is enabled by a simple rule: every time a unit adapts it may use its \textit{near links} to potentially trigger other units to adapt as well. 
 Thus one adaptation may trigger another, which in turn may trigger another - in a domino-like fashion. 
 
 We suggest that the scheme described in the previous paragraph can be used as a general method to train topographically preserving feature maps, in an asynchronous fashion. 
 We present a simple implementation of this scheme, and provide empirical evidence in support of this argument. 
  Our key contribution lies in demonstrating a highly scalable asynchronous training algorithm for topographic maps. 

 As a first investigation of this scheme, and to demonstrate the feasibility of this claim, we present a basic implementation of the heuristic search, the rules governing cascading, 
 and the rules for creating near and far links. In section \ref{dsom} we use MNIST to explore the behavior of the proposed algorithm, which results in a set of suggested hyper-parameters. 
 To demonstrate the applicability and robustness of the proposed algorithm hyper-parameters configuration, we use the algorithm for classification on  $4$ different datasets (including MNIST) in section \ref{experiments}. 
 We are also able to derive the computational complexity of presented algorithm, under the proposed hyper-parameter configuration. 
 We compare the resulting classification performance to values reported in the literature for a SOM, and find that the two methods perform comparably.

\color{black}

\section{Algorithm} \label{dsom}


Each of the $N$ units is given a position in two spaces: the unit space and the sample space. 
In the unit space, units are arranged in a regular lattice, $\left\{ 0, .. \sqrt{N} \right\}^2$. 
Unit's $j$ position in the sample space is given by the weight vector $w_j$. 
As part of the cascading mechanics, each unit is given a counter $c_j$ (initialized with $0$), and all units share a common cascading threshold $\theta$. 
Through training, we seek to obtain weights $w_j$ such that: \textbf{i)} the topologies of the units in the two spaces (sample and unit space) are similar, and \textbf{ii}) the weights accurately approximate the distribution of the samples in sample space.

\textbf{Links} are drawn based on the Manhattan distance of units $j,k$ in the unit space $D_{jk}$. 
\begin{itemize}
	\item{
		\textbf{Near links} are used for the heuristic search and adaptation. These links are drawn if $D_{jk} \leq 1$, forming a square lattice. The units linked to unit $j$ through near links are the \textit{near neighbors} of $j$, denoted by $\mathcal{N}_j$.  
	}
	\item{
		\textbf{
			Far links} are only used for the heuristic search. 
		They are drawn probabilistically, by connecting each unit $j$ to a fixed number $\phi$ of its peers, with a probability\footnote{Such connection probability parameterizations (distance based power-laws) are known to perform optimally in certain heuristic search problems involving spatial graphs \cite{kleinberg2000navigation} - but different schemes may be consider in future works.} $\sim D_{jk}^{-1}$. 
		Units connected to unit $j$ through far links are the \textit{far neighbors}, denoted by $\mathcal{F}_j$. Details on the choice of $\phi$ are found in the respective section.
	}
\end{itemize}

Training the map involves two processes: the distributed heuristic search for the GMU, and the cascade-driven adaptation. Adaptation relies solely on the near links, while the heuristic search utilises both near and far links.

\subsection{The distributed heuristic search} \label{thesearch}
The search aims to determine the GMU $j^*$, in a fashion akin to a relay-race: the sample is passed between units, in an effort to find a $j^*$ that minimizes:
\begin{equation}
q_{ij^*} = |w_{j^*} - s_i|. 
\end{equation}
The search involves two phases. 
Starting from a random unit $j$ (who we also set as $j^*$), we take the following steps:
\begin{enumerate}
	\item{
		\textbf{Random exploration}, the index of the unit holding the sample $j$ is updated by picking a far neighbor or $j$ uniformly at random. 
		The sample is sent there, and if $q_{ij} < q_{ij^*}$  we update the GMU: $j^* \leftarrow j$. 
	}
	\item{We repeat step 1, for a total of $e$ iterations (where $e$ is a user defined parameter). We then polish the best know solution $j^*$ through a greedy search. }
	\item{
		\textbf{Greedy exploitation} The sample is compared to the near and far neighbors of $j^*$, and determine $k^*$:
		\begin{equation}
		k^* \leftarrow \underset{k \in \mathcal{N}_j}{\text{argmin}} ~~ q_{ki} 
		\end{equation}
	}
	\item{If we get $q_{ik^*} < q_{ij^*}$ the GMU index is updated $j^* \leftarrow k^*$ and step 3 is repeated. Otherwise, we the search is terminated and $j^*$ is the GMU.}
\end{enumerate}

\begin{figure}
	\centering
	\includegraphics[scale=0.12]{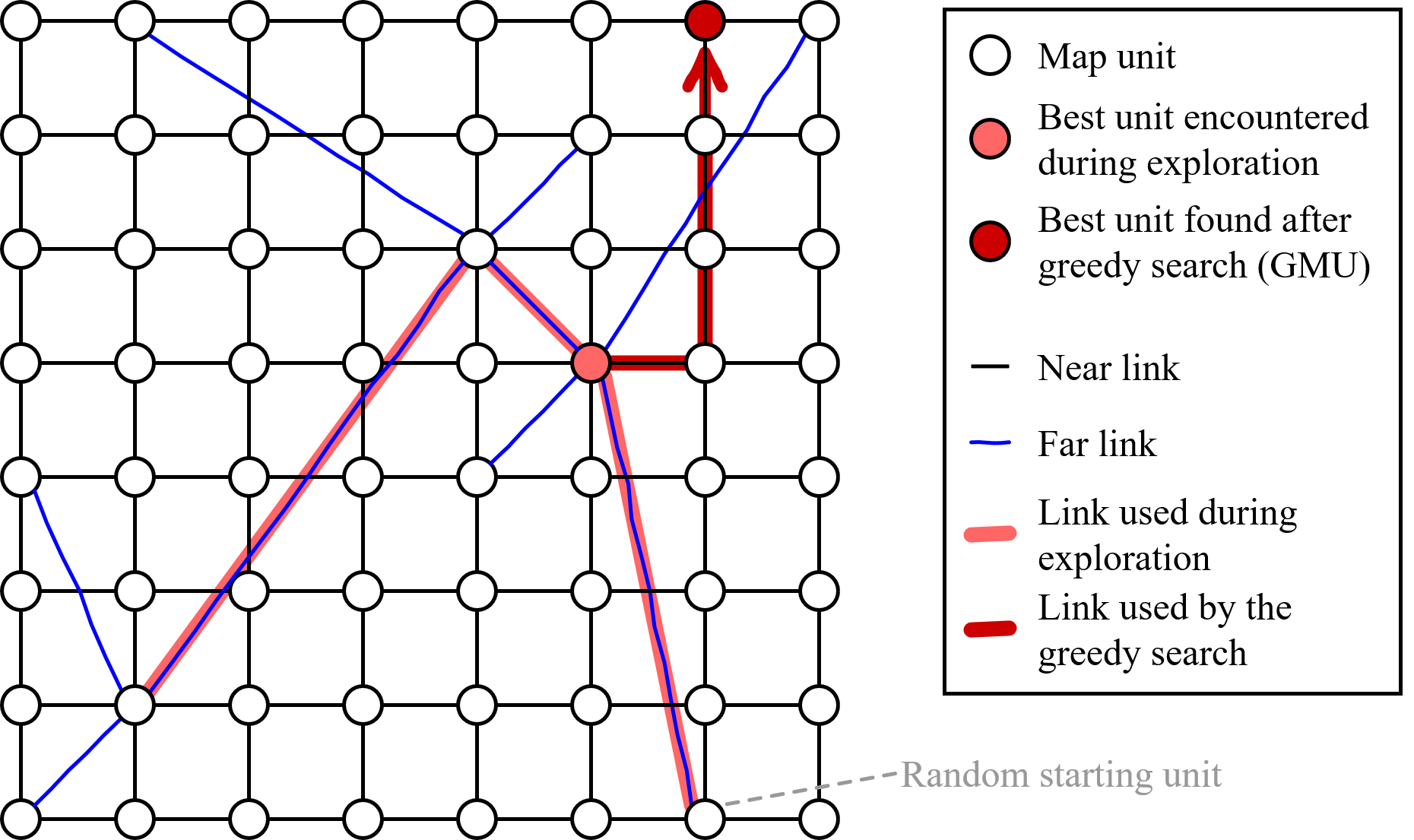}
	\caption{
		Visualizing the proposed heuristic search. 
		Units are linked to near and far neighbors (black and blue lines). 
		Each sample hops over the links, in an effort to find a its best-matching unit.  
		The search has two phases: random exploration over the far links (pink highlight), and a greedy search over near links (red highlight). 
		Far links allow for effective exploration (details in section \ref{thesearch}). 
		We depict a small fraction of the far links. }\label{dSearch}
\end{figure}
The number of exploration iterations $e$ (see step 2) can be used to adjust the accuracy of the search. 

To quantify the performance of this search, we may check if the unit identified as the GMU is in fact the best matching unit (BMU) - that is, the global optimum of the problem ${argmin}_j ~~ q_{j}(i)$. 
If a search results in a GMU that does not coincide with the BMU we say that the search has failed. 
In the following subsection, the performance of the search is quantified by calculating the fraction number of failed searches towards the end of training. 
We refer to this quantity as \textit{search error}, denoted by $F$.

The pathology arising from an inaccurate search, along with an empirical investigation on how $e$ impacts search accuracy for different map sizes $N$ can be found in the next section (see subsection \ref{exp:search}). 
Additionally, the search algorithm is presented in pseudo-code form in Algorithm \ref{psudocode} and visualized in Figure \ref{dSearch}.

Note that the distant connections in $\mathcal{F}_j$ allow for long jumps in the unit space prior to (and during) the greedy search - helping the algorithm escape from local minima. 
Sufficiently increasing the number of far connections per unit $\phi$ allows for the random search process to quickly diffuse over the map, in $\mathcal{O}(\log(N))$ iterations \cite{martel2004analyzing} - ensuring the scalability of the search with respect to map size.  
This effect is well known in graph theory as the \textit{small-world} effect \cite{milgram1967small}, and it can be achieved with a relatively small number of $\phi$ \cite{barriere2001efficient}. 
This search algorithm suffices for demonstration purposes, but more sophisticated schemes may be considered in the future. 

\subsection{Cascade-driven adaptation}\label{cascading} 
Adaptation occurs once the GMU $j^*$ is determined. Let $l_s$ by a user defined constant learning rate, the weight vector is updated by:
\begin{equation}
w_{j^*} \leftarrow w_{j^*} + l_s (s_i-w_{j^*})
\end{equation} 
Additionally, the cascading counter of the GMU is increased (by applying $c_{j^*} \leftarrow c_{j^*}+1$) with a probability $p_i$. 
The exact parametrization of $p_i$ (dubbed the cascading probability) will be discussed later in this section.
Afterwards, cascading adaptation may take place through the following rules:
\begin{enumerate}
	\item{\textbf{Firing:} 
		If a cascading counter update yields $c_j > \theta$, the unit is said to \textit{fire}: it resets $c_j \leftarrow 0 $, and broadcasts $w_j$ to all its near neighbors.}
	\item{
		\textbf{Cascading adaptation:}
		Upon receiving a weight vector $w_k$, unit $j$ \textit{adapts} its weight own vector:
		\begin{equation} 
		w_j \leftarrow w_j + l_c(i) (w_j - w_k)
		\end{equation}
		where $i$ is the index of the last sample, and $l_c(i)$ is a learning rate which depends on the training index $i$.}
	\item{\textbf{Drive:} Following every adaptation of $w_j$ we apply $c_j \leftarrow c_j + 1 $ with a probability of $p_i$. }
\end{enumerate}
The cascading adaptation process is also described in pseudo-code from in Algorithm \ref{psudocode}. The magnitude of the cascading event associated with sample $i$ is measured by the \textit{cascade size} $a_i$ - which is calculated by counting the number of firing incidents after all firing has ceased. 
In the following analysis, we commonly refer to the fractional cascade size, which is defined as $A_i = a_i/N$.

Each firing incident results in a unit attracting its near neighbors in the sample space. 
This increases the topological order of the map - but may also reduce the similarity of the weight vectors to the samples.  
The attraction between weights can be controlled via the learning rate $l_c(i)$, and $p_i$. 
In an effort to globally order the map, we allow for large scale cascading (of scale $\mathcal{O}(N)$) and high $l_c$, during early training.  
Then, we gradually reduce both $l_c, p_i$ over the course of training, allowing the weights to better approximate samples. 

The cascading learning rate follows:
\begin{align}
l_c(i) &=  \left( 1 + tanh \left( \frac{c_{o}- i/i_{max}}{c_s} \right) \right) \bigg/  2  \label{eq:lc}  
\end{align}
Where $i_{max}$ is the total number of training iterations, and $c_o,c_s$ are user defined constants.  
Equation \eqref{eq:lc} results in $l_c(i)$ following a smoothly decreasing slope as training progresses - while ensuring $l_c(i)\in (0,1), ~ \forall i$. 
The \textit{offset} parameter $c_{o}$ controls how late into training $l_c$ reaches the value $0.5$ - e.g. setting $c_o=0$ or $c_o=1$ results in $l_c(0)=0.5$ and $l_c(i_{max})=0.5$ respectively. 
Parameter $c_s$ control the slope of decrease - with $c_s=0$ resulting in the instantaneous decrease from 1 to 0, and $c_s \rightarrow \inf$ resulting in constant $l_c$.

\begin{algorithm}[htp!]
	
	e :	Number of Explorations\\
	l\_s : Learning Rate for Samples\\
	$\theta$ : Cascading Threshold\\
	$i\_max$ : Total Number of Training Samples \\

	\setcounter{AlgoLine}{0}
	\DontPrintSemicolon
	\SetKwFunction{FMain}{}
	\SetKwProg{TM}{Function TrainMap}{:}{}
	\TM{\FMain{}}{
		ConnectNearNeighbors() \\
		ConnectFarNeighbors() \\
		\For{i: 1 to i\_max}{	
			sample = getRandomSample()\\
			bmu = HeuristicSearch(sample)\\			
			AdjustWeight( bmu, sample, l\_s)\\	
			IncrementGrain(bmu,i)\\
			\If {getGrains(bmu) $>$= $\theta$}	
			{ 
				Cascading(bmu, i)
			}	
			
		}
		
	}
	\;\\
	
	\setcounter{AlgoLine}{0}
	\DontPrintSemicolon
	\SetKwProg{Hc}{\textbf{async} Function HeuristicSearch}{:}{}
	\Hc{\FMain{$sample$}}{
		
		bmu = getRandomUnit()\\	
		rnd\_neighbor = bmu\\		
		\For{1 to e}{			
			rnd\_neighbor = getRandomFarNeighbor( rnd\_neighbor )\\			
			temp = getDistance(rnd\_neighbor, sample)\\			
			\If{temp$<$ min\_dist}	
			{	
				bmu = rnd\_neighbor\\	
				min\_dist = temp	
			}
			
		}			
		temp =  getMinDistNeighbor(bmu, sample)\\
		\While{temp $<$ min\_dist}	
		{	
			bmu = neighbor\\	
			min\_dist = temp\\			
			neighbor = getMinDistNeighbor(bmu, sample)\\	
			temp= getDistance(neighbor , sample)
		}
		\KwRet bmu \; 
	}
	
	\;\\

	\setcounter{AlgoLine}{0}
	\SetKwProg{Fn}{\textbf{async} Function Cascading}{:}{}
	\Fn{\FMain{$unit$, $i$}}{
		EmptyGrains(unit)\\
		\For {each near neighbor of unit}
		{
			
			AdjustWeight( neighbor, getWeight(unit), getCascadingLearningRate(i))\\
			\If  {rnd $<$ getCascadingProbability(i)}{ addGrain(unit)}
			\If {getGrains(neighbor) $>$= $\theta$}	
			{
				Cascading(neighbor, i)
			}	
		}
		\KwRet\;
	}

	\caption{Pseudocode for the proposed approach to training topographic feature maps asynchronously.}
	\label{psudocode}
\end{algorithm}

\textbf{Manipulating cascade sizes} 
To control the macroscopic cascading behavior we parametrize $p_i$ by relying on methods from statistical physics. 
The dependence between the statistical behavior of $a_i$ and $p_i$ (that is, the probability distribution $\mathbb{P}(a_i=C; p_i)$) can be analytically understood under four assumptions:
\begin{itemize}
	\item{samples are spread homogeneously across the unit space throughout training,}
	\item{the value of $p_i$ changes slowly during training (known as the adiabatic approximation),}
	\item{the time that passes between presenting the map with samples is enough for cascading to cease, }
	\item{the number of near links equals the threshold $\theta = |\mathcal{N}_j|$.}
\end{itemize}
Based on these assumptions - and for $p_i=1$ - cascading can be exactly mapped to the BTW sandpile model \cite{bak1988self}.
For $p_i < 0$ cascading can be mapped to a non-conservative variant of the well established sandpile model where dissipation occurs with probability $1-p_i$ \cite{vespignani1998driving, malcai2006dissipative}.
In this (dissipative) variant, cascade sizes are known to follow a power law distribution - truncated exponentially at a characteristic fractional size $\bar{\chi}$.
In dissipative sandpile models, it generally holds $\bar{\chi} \sim d^{-s}  $ where $d$ is the rate of dissipation, and $s$ a critical exponent which depends on the specific rules governing cascading \cite{vespignani1998driving, malcai2006dissipative}. 
In our case, $s=1$ \cite{vespignani1998driving}, and dissipation is given by $d \sim 1-p_i$.
Thus, we can manipulate the characteristic cascade size $\bar{a}_i$ by adjusting $p_i$. 

In order to study the behavior of the algorithm under varying map sizes $N$, it is instructive to parametrize $p_i$ in a way that results in \textit{scale invariant dynamics}: that is, the relative size of cascades $a_i / N$ is independent of $N$. 
For $N$ sufficiently large, we can achieve scale invariance by the following parametrization:
\begin{equation}
p_i = \left( 1 - \frac{1}{\sqrt{c_m N}} \right) \left( 1- \frac{i}{i_{max}} \right)^{\frac{c_d}{N}}   \label{eq:pi} 
\end{equation}
This relation results in a high starting value of $\bar{a_i}$ which reduces over time. 
The constant $1/N << c_m < 1 $ controls the characteristic size $\bar{a_i}$ in early training, while $c_d \in (0, \infty )$ controls the rate at which $\bar{a_i}$ shrinks over time. 
The above behavior is independent of the system size $N$.

\section{Experimental Analysis} \label{experiments}

In this section we explore the impact of the model's hyper-parameter on the quality of the resulting map. 
Our analysis is focused on the hyper-parameters that govern the two novel components of the presented algorithm: the cascading adaptation mechanism and the distributed heuristic search.

\paragraph{Default configuration} Unless otherwise stated, for all experiments of this section, we will be considering the MNIST dataset and a map of 900 units, with the following configuration:
The number of far connections $\phi$ is set to 20 (which ensures that the network is densely connected).
The learning rate parameters are set to $l_s = 0.05$ , $c_o = 0.5$ and $c_s = 0.5$. Cascading parameters are set to $c_m = 0.1$ and $c_d = 100$. Exploration iterations are set to $e = 3N$.  
The number of training iterations are proportional to the number of the units (as is common practice for the closely related algorithm SOM \cite{massivesom}). 
The number of epochs are adjust so that the number of training samples is $i_{max} \approx 600N$. \label{default_hyper}

\paragraph{Measuring map quality} Map quality can be assessed in two ways, using two different, well-established metrics. Quantization error quantifies how well the weight vectors describe the distribution of the samples in the sample space, while Topological error quantifies the topological order the map (on a local scale) \cite{topdistoration_LR}. 
The quantisation and topological error of a map are denoted by $Q$ and $T$ respectively.  

\subsection{Heuristic search accuracy}\label{exp:search}
To explore the impact of the exploitative search iterations parameter $e$ on the resulting map, we train maps using an increasing number of search iterations: $e \in \{0.01 N, 0.05N, 0.1N, 0.2 N, 0.3 N \ldots  5N \}$ and compare the resulting search accuracies. To quantify search accuracy, we focus on the last $1000$ training iterations, and calculate as the fraction of times that the GMU did not coincide with the BMU. This fraction is denoted by $F$.  

\begin{figure}[h!]
	\centering
	\includegraphics[width=0.5\textwidth]{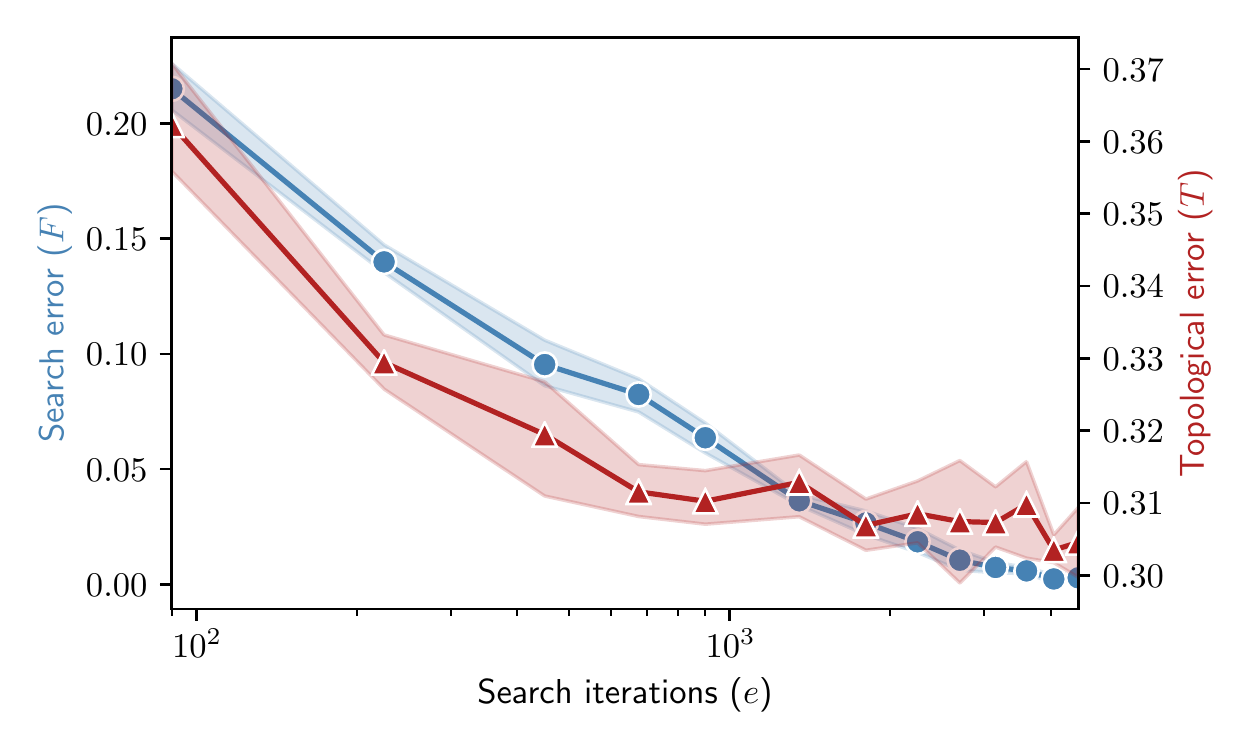}
	\caption{Increasing the exploitative search iterations $e$ asymptotically reduces search error (blue line) and topological error (red line). 
		Increasing search accuracy has a diminishing effect on map quality. 
		Confidence intervals are placed at one $std$, and calculated over $5$ runs. 
		For the experiment's configuration see section \ref{default_hyper} - second paragraph.}
	\label{fmbusVSE}
\end{figure}

The results of this experiment can be seen in figure \ref{fmbusVSE}, which reveals that the accuracy of the heuristic search increases exponentially over the range of considered $e$. 
The figure also depicts the topological error of the map as a function of $e$, revealing that further increasing the search accuracy offers diminishing improvements to topological quality.
Based on these results, we will be setting the hyper-parameter $e=3N$ for all remaining experiments (as $3N$ is the smallest value of $e$ the results in search accuracy $>99\%$).
For $e=3N$, the scalability of the search with respect to $N$ is established in a subsequent experiment (see subsection \ref{exp:scalability}) which demonstrates that the map quality improves as $N$ increases. 
More thorough analyze can be found in the Appendix.

\subsection{Cascading adaptation}\label{exp:cascading}

The cascading adaptation mechanism relies on two hyper-parameters, one associated with the average cascade size during early training, and one with the rate at which cascade sizes decrease over the training duration ($c_m, c_d$ respectively). 

\begin{figure}[h!]
	\centering
	\includegraphics[width=0.5\textwidth]{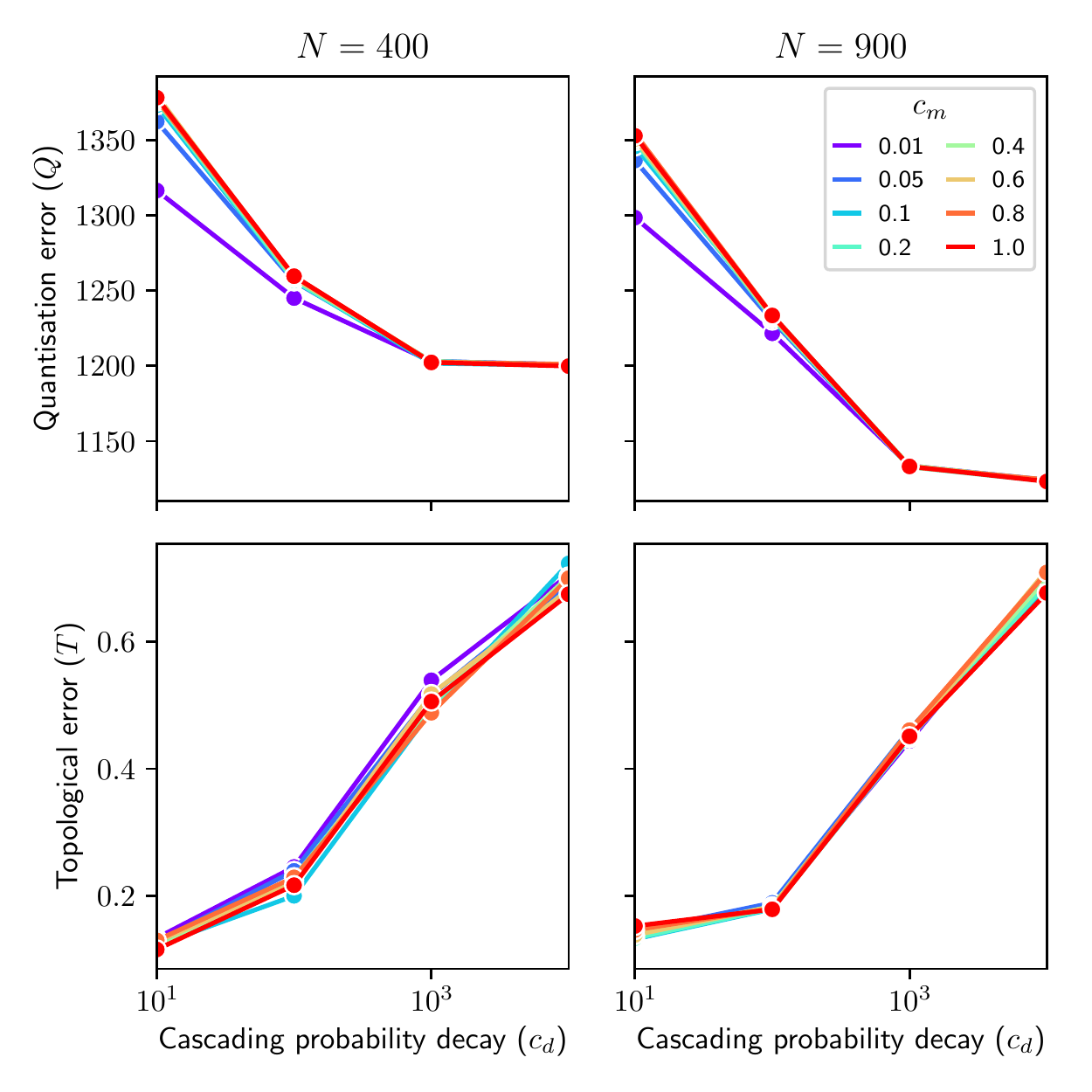}
	\caption{
		A sparse grid search of the cascading parameters $c_d, c_m$ reveals that $c_m$ has minimal effect over the map's quantization and topological errors ($Q$ and $T$). 
		The grid search is repeated for two map sizes $N$: $400, 900$ (left column and right column respectively). 
		For the experiment's configuration see section \ref{default_hyper} - second paragraph.
	}\label{CascadeSearch}
\end{figure}

To assess the impact of these two parameters, we train maps configured over a sparse grid of $(c_m, c_d)$, with $c_m \in \{ 0.01, 0.05 0.1 \ldots 1 \}$ and $c_d \in \{ 10, 10^2, 10^3, 10^4 \}$.The topological and quantization errors of the resulting maps are shown in figure \ref{CascadeSearch} - for two map sizes. 
Figure \ref{CascadeSearch} reveals that quantization and topological error are not sensitive to $c_m$. We can therefore select a small $c_m$, which result in smaller cascades and reducing computation.
However, $c_m$ must be $>>1/N$ (see equation \eqref{eq:pi}) - and thus we select $c_m=0.1$ for the remaining experiments as a compromise. 

\begin{figure}[h!]
	\centering
	\includegraphics[width=0.5\textwidth]{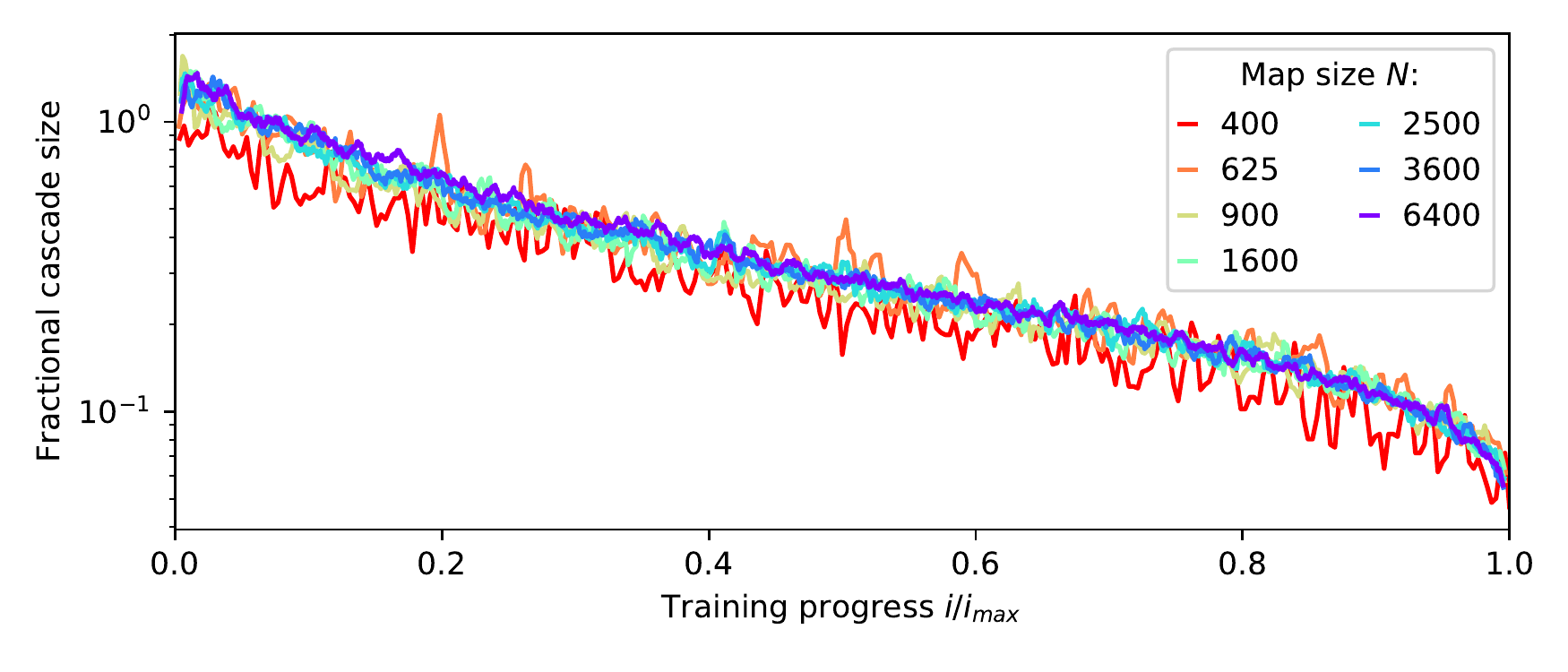}
	\caption{
		Increasing map size $N$ does not significantly affect fractional cascade sizes $A_i$. 
		We track the upper $1000$th quantile of $A_i$ during training (over a rolling window of width $i_{max}/100$) - for increasing map sizes $N$. 
		Tracked trajectories collapse to a single time-line, indicating the $N$ does not affect fractional cascade sizes. 
		For the experiment's configuration see section \ref{default_hyper} - second paragraph.
	}\label{CascadingScaleInvariance}
\end{figure}
\begin{figure}[h!]
	\centering
	\includegraphics[width=0.5\textwidth]{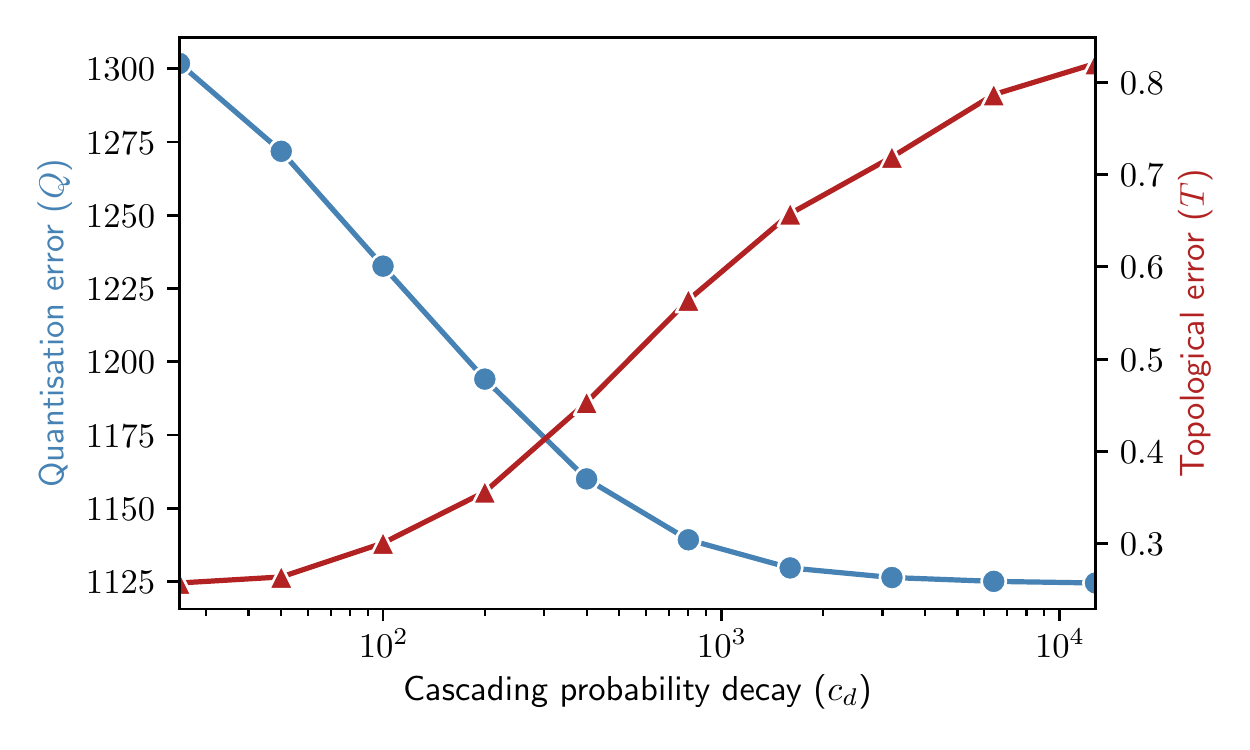}
	\caption{
		Inreasing the cascading decay parameter $c_d$ has a mixed effect on map quality: quantisation error $Q$ decreases at the expense of increasing topological error $T$. 
		For the experiment's configuration see section \ref{default_hyper} - second paragraph.
	}\label{QTDecay}
\end{figure}

To more closely examine the impact of the other cascading hyper-parameter $c_d$ on the algorithm's performance, we train maps with increasing $c_d$, and measure their quantization nd topological errors. 
We demonstrate results in figure \ref{QTDecay}, which reveals that $c_d$ has a mixed impact on the map quality: increasing $c_d$ may reduce quantization error at the expense of increasing topological error. 
Thus, the selection of an appropriate value for $c_d$ may depend on the intended use of the resultant map. 
We select $c_d = 100$ and $c_d=1000$ in order to emphasis low topological, and quantization error - respectively. 

Cascading is controlled via the cascading probability $p_i$, which - as stated in subsection \ref{cascading} - is parametrized in such a way that the cascading behavior is naturally adjusted to the system size. 
Specifically, the fractional size of cascades $A_i$ is controlled by $c_m, c_d$ and decoupled from the map's absolute size $N$. 
This statement is verified empirically, by training maps of increases sizes $N \in \{ 100, 225, 400, 625, 900,1600, 2500,3600,6400   \}$ and comparing the resulting cascading activity. 
To compare the cascading activity between maps of different sizes we i) take a rolling window of width $i_{max}/100$, and ii) take the mean of fractional cascade sizes $A_i$ that lie in the top $1000$th quantile. 

The results of this analysis are depicted in figure \ref{CascadingScaleInvariance}, which shows that the lines plotted for all map sizes coincide - demonstrating that by using the presented parametrization of $p_i$, the fractional cascade size $A_i$ does not strongly depend on the system size - for $N>400$.

\subsection{Scalability} \label{exp:scalability}
\begin{figure}[h!]
	\centering
	\includegraphics[width=0.5\textwidth]{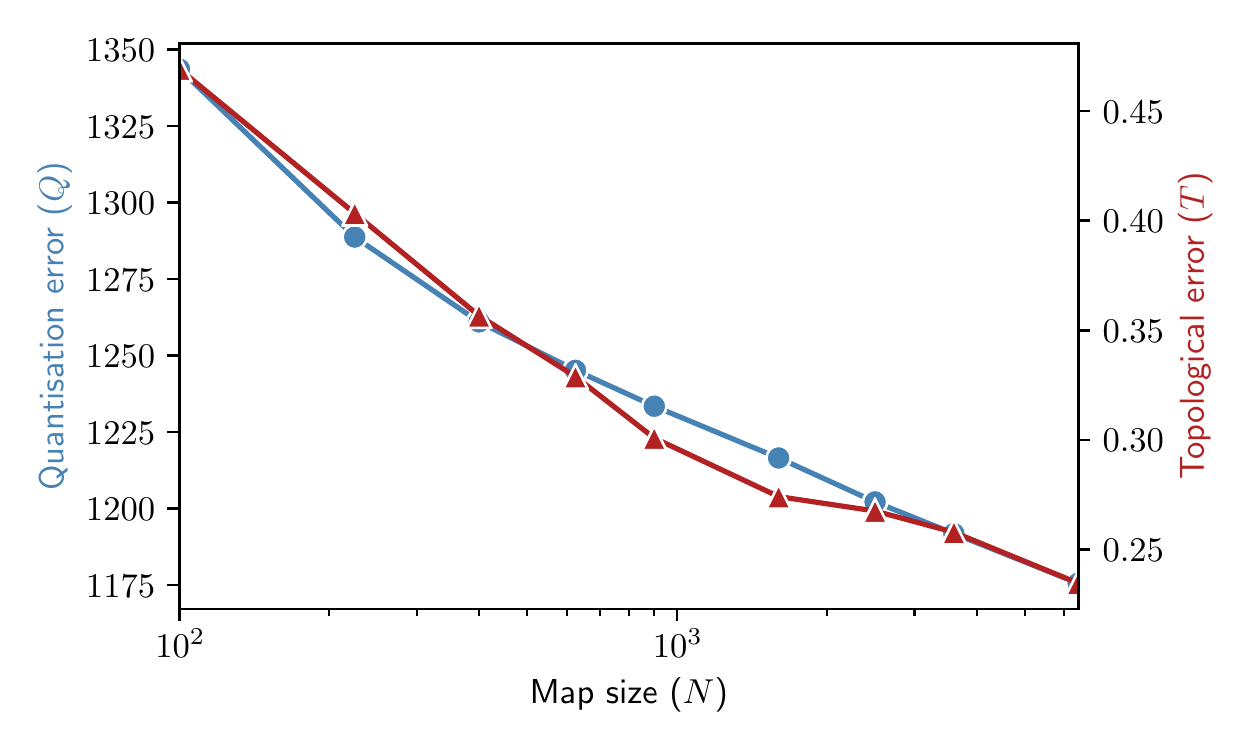}
	\caption{
		Increasing map size $N$ results in lower topological and quantization errors ($Q$ and $T$) - demonstrating that the quality of maps increases with system size - assuming large $N$. 
		For the experiment's configuration see section \ref{default_hyper} - second paragraph.
	}\label{QTvsN}
\end{figure}

To investigate the scalability of the algorithm, we train maps of increasing sizes $N \in \{ 100, 225, 400, 625, 900,1600, 2500,3600,6400 \}$ - for $c_d=100, c_m=0.1, e=3N$. 
We measure the quality of each trained map (that is quantization error, topological error, search accuracy (defined in the section \ref{thesearch}).
The results are presented in the figure \ref{QTvsN}, and indicate the algorithm's scalability in terms of performance: both quantization and topological error reduce exponentially over the considered range of $N$.
This observation reveals that hyper-parameter values are not sensitive to system size - which is highly instructive for tuning (i.e. a configuration that works on a small map can also be used for a larger map). 
We attribute this convenient property to the scale-invariant cascading parametrization, and the scalability of the heuristic search. 

\begin{table}[ht!]\label{data_T}
	\centering
	\begin{tabular}{|c|c|c|c|} 
	
		\hline
		Dataset & Classes &       Features          &         Samples        \\
		\hline 
		FMNIST & 10 &  784  &  59999      \\
		 &  &    &  10000      \\
		  Letters & 26 &  16  &  15000      \\
		 &  &    &  5000      \\
		 MNIST & 10 &  784  &  59999      \\
		 &  &    &  10000      \\
		  Sat. images & 6 &  36  &  4435      \\
		 &  &    &  2000      \\
		\bottomrule
	\end{tabular}
	\caption{Description of the datasets used to evaluate the algorithm's classification performance and dataset-sensitivity. Sources of the datasets can be found in \cite{melka2017efficient}}
	
\end{table}

\begin{table*}[htp!]
	\centering
	
	\begin{tabular}{|cc|c|c|c|c|} 
		\hline
		&&\multicolumn{2}{c|}{AFM}&\multicolumn{2}{c|}{Sparse-SOM}\\
		\hline
		 Datasets   &  &    Precision          &         Recall  & Precision   & Recall    \\
		\hline
		FMNIST & test &  $78.0 \pm 0.6$ &  $76.8 \pm 0.6$  &  &  \\
		& training &  $78.0 \pm 0.3$ &  $76.8 \pm 0.4$  &  &  \\
		Letter & test & \boldmath{$83.3 \pm 0.2$} & \boldmath{$82.8 \pm 0.3$} &$81.5 \pm 0.5$   &$81.1 \pm 0.5$\\
		& training &  \boldmath{$86.0 \pm 0.2$}&  \boldmath{$85.7 \pm 0.3$} & $83.7 \pm 0.3$ & $83.7 \pm 0.3$ \\
		MNIST & test &  $93.3 \pm 0.1$ &  $93.2 \pm 0.2$ & \boldmath{$93.4 \pm 0.2$} & \boldmath{$93.4 \pm 0.2$} \\
		& training &  $93.1 \pm 0.2$ &  $93.1 \pm 0.2$ &\boldmath{$93.5 \pm 0.2$} & \boldmath{$93.5 \pm 0.2$} \\
		SatImage & test &  \boldmath{$87.8 \pm 0.4$} &  \boldmath{$87.6 \pm 0.6$} & $87.6 \pm 0.3$ & $85.3 \pm 0.4$ \\
		& training &  $91.7 \pm 0.2$ &  $91.5 \pm 0.2$ & \boldmath{$92.3 \pm 0.4$} & \boldmath{$92.4 \pm 0.3$} \\
		\bottomrule
	\end{tabular}
\caption{
	Comparing the classification performance of the proposed feature map (AFM) to that of the self organising map (SOM) - using values from literature \cite{melka2017efficient}. 
	The average and the standard deviations of Recall and precision are calculates for five runs of AFM, for multiple values. 
	The results are comparable with the performance of SOM. 
	We use a $34 \times 34$ AFM with $c_d=1000$ - for the remaining hyper-parameter values see section \ref{default_hyper} - second paragraph.}

	\label{class_T}
\end{table*}

\subsection{Applicability} \label{applicability}
In this subsection we demonstrate that the presented algorithm can perform comparably to its closest literature variant (SOM) over a diverse set of datasets. 
Through this demonstration, we also verify that the proposed hyper-parameters are robust to over a wide range of different datasets. 
That being said, we point out that the goal of this study is not to outperform the state of art, but to demonstrate a novel approach to training feature maps in an asynchronous fashion. 
Optimizing the algorithm's performance (by further calibrating hyper-parameters, or increasing $N$) can be the subject of future studies. 

In order to more easily compare the results of the algorithm to the literature, we use a map of $34 \times 34$ units - resulting in $N=1156$. 
We also set $c_d=1000$ which results on lower quantization error (see subsection \ref{exp:cascading}). 

\paragraph{Classification}
In this section we use the presented asynchronously trained feature map (AFM) in a classification task. 
We purposefully use a simple classification scheme - nearly identical to the one used in \cite{melka2017efficient}. 
We do so for the sake of succinctness, and to allow for the meaningful comparison between our work and the results presented in \cite{melka2017efficient}. 
\begin{enumerate}
	\item{}
	Each map unit $j$ is assigned with a class $y_j$ , which represents the sample class the that unit is most similar too. 
	These classes are assigned to units after training has finished. 
	\item{}
	Let $Y_i$ denote the label of sample $i$.  
	We first find the most similar sample to $w_j$: 
	\begin{equation}
	i^*(j) =  \underset{i}{\text{argmin}} ~ |w_j - s_i|
	\end{equation}
	Then, unit $j$ is given the class of sample $i^*$: $y_j \leftarrow Y_{i^*}$. 
	\item{}
	Finally, to predict the label of a sample, we find the sample's BMU and the return the class of the BMU.
\end{enumerate}

Table \ref{class_T} summaries the classification performance of the algorithm, by presenting the mean and standard deviation over $5$ runs. 
Additionally, the table presents the performance metrics of an SOM of $1200$ units - as they were presented in \cite{melka2017efficient}. 
Since the distance in performance in relatively small, we may consider that the presented algorithm performs comparably to an SOM of comparable size.

\begin{figure}[h!]
	\centering
	\includegraphics[width=0.5\textwidth]{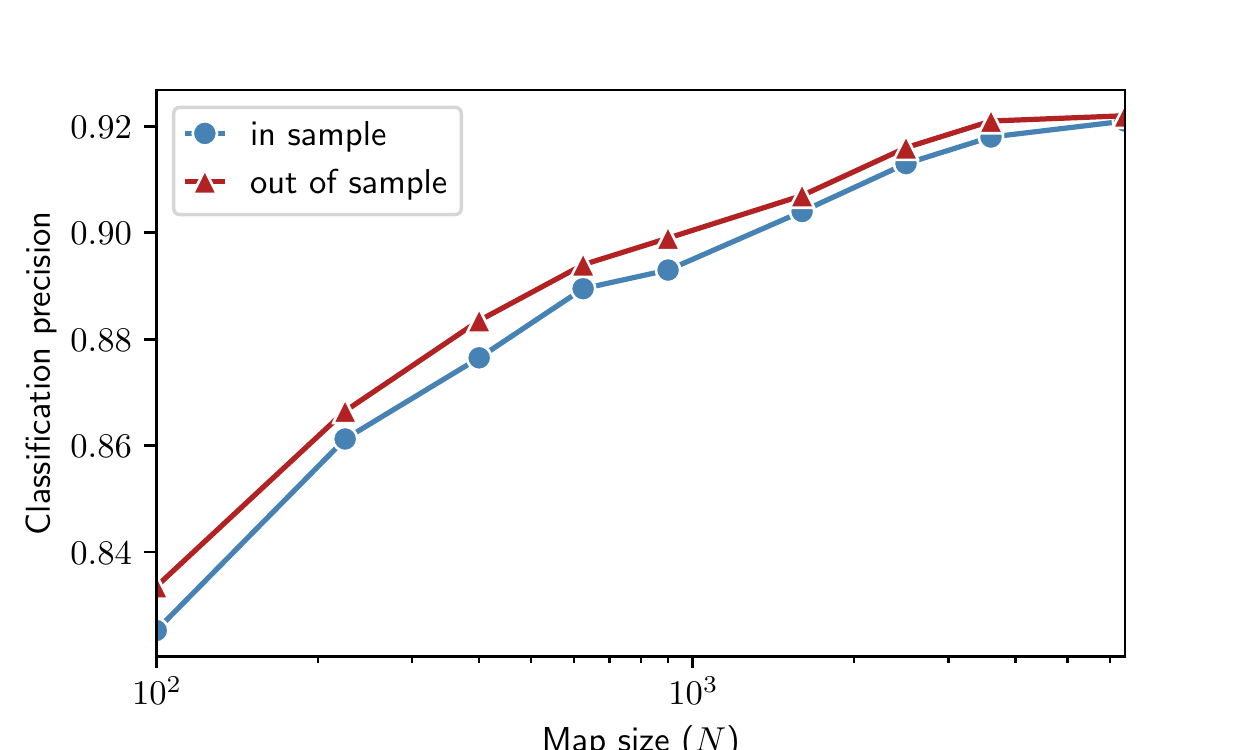}
	\caption{
		When used in classification, the precision of the proposed map increases with map size $N$ (both in and out of sample). 
		The classification scheme used is presented in subsection \ref{applicability}. 
		For the experiment's configuration see section \ref{default_hyper} - second paragraph.}
	\label{pic:classvsN}
\end{figure}

The figure \ref{pic:classvsN} demonstrates that classification precision increases with increasing system size.

Finally, table \ref{t:behavior_T} allows us to assess the dependence of the heuristic search algorithm and the cascading mechanism on the structure of the data. 
Specifically, table \ref{t:behavior_T} presents the number of weight update operations that took place after each sample - averaged throughout the entire sample period. 
The maximum relative pairwise distance between the reported number is approximately $1.5\%$, which implies that the intensity of cascading was comparable across all simulations. 
Table \ref{t:behavior_T} also presents the search accuracy towards the end of training of each map - averaged for each dataset (for the definition of search accuracy see subsection \ref{thesearch}). 
The search error varies between datasets, but generally remains low - with its maximum at $2.74\%$ for MNIST. 
The variability of search accuracy across datasets may be attributed to the differences in the dimensionality of the datasets (since higher dimensional sample spaces may be harder to search).

\subsection{Computational complexity}\label{complexity}

The previous experiments demonstrate that the presented algorithm can perform comparably to SOM for a hyper-parameter configuration that obeys: $e \sim N, p_i \leq 1, i_{max} \sim N$). 
This insight allows us to derive an upper bound of the algorithm's complexity, as follows.

For each of the $i_{max} \sim \mathcal{O}(N)$ training sample, we do one exploitative search of $e \sim \mathcal{O}(N)$ iterations, followed by a greedy search of $g_i$ iterations. Since no unit can be visited twice during the greedy search it is $g_i \leq N$ and thus $g_i \sim  \mathcal{O}(N)$. 
Finally, after the search is over we proceed with cascading. 
Cascading sizes are maximized for $p_i=1$ \cite{vespignani1998driving} - for which the presented model can be exactly mapped to the sandpile model from statistical mechanics \cite{bak1988self}.
It is well known that cascade sizes for the sandpile model scale like $N$ \cite{bak1988self}. 
Since in the presented algorithm $p_i \leq 1$, the linear scaling of sandpile is a upper bound for our model, $\bar{a}_i \sim N$ (the overbar $\bar{\cdot}$ denotes an upper bound). 

\begin{table}[htp]
	
	\setlength{\tabcolsep}{1pt}
	\centering
	\begin{tabular}{|c|c|c|c|}
		\hline
		Dataset & \makecell[c]{Max fractional \\  cascade} & \makecell[c]{Number of weight\\ updates / Sample} & Search error \\
		\hline
		FMNIST&$0.75 \pm 0.11$ &$3.16 \pm 0.01$ &$2.46 \% \pm 0.26\%$ \\
		Letter&$0.76 \pm 0.15$ &$3.18 \pm 0.01$ &$1.62\% \pm 0.28\%$ \\
		MNIST&$0.87 \pm 0.11$ &$3.18 \pm 0.01$ &$2.74\% \pm 0.74\%$ \\
		SatImage&$0.78 \pm 0.17$ &$3.21 \pm 0.02$ &$1.62\% \pm 0.56\%$ \\
		\hline
	\end{tabular}
	\caption{Largest fractional cascade $\max {A_i}$, the average cascade size $\text{avg} (a_i)$ and the average search error $F$ for different dataset. Values are comparable across datasets, implying that algorithm behavior is not sensitive to the specificities of datasets.}
	\label{t:behavior_T}
\end{table}

Based on the previous paragraph we can derive the computational complexity of the presented hyper-parameter parametrization ($e \sim N, p_i \leq 1, i_{max} \sim N$):
\begin{equation}
\mathcal{O}(i_{max} ~ (e + \bar{g}_i + \bar{a}_i)) = \mathcal{O} \big(N (N+N+N) \big) = \mathcal{O} (N^2)
\end{equation}
Therefore the presented parametrization has a computational complexity of $N^2$ - which is in the same class as that of its closest literature variant (the SOM \cite{massivemap}).


\section{Discussion}
Currently, all state of art training algorithms for topographic maps (and their variants) rely on densely connected neurons. This is in spite of the scalability advantages that sparse topologies have to offer \cite{mocanu2018scalable}. To boost the computational performance of topographic maps, the state of art relies on distributed map-reduce platform  \cite{parallelSOM,parallelcudaSOM}. However such solutions require synchronization and central entity - and are thus vulnerable to the limitations of the communication channel (i.e. delays, bandwidth, reliability) and slow workers \cite{mnih2016asynchronous,li2019generalized,fernando2017pathnet}. 

These limitations can be overcome by enabling the asynchronous execution of the training process. Broadly speaking, the challenge in enabling the asynchronous execution of an algorithm lies in ensuring a sufficiently low level of dependencies between the algorithm's segments - a.k.a. loose coupling. To achieve that, we propose a topographic map training algorithm distributes all computation to a swarm of autonomous units. 
Each unit is only aware of a few pre-defined neighbors, with which it interacts infrequently - resulting in loose coupling. 

We investigate the behavior of this scheme empirically on MNIST, and prescribe a concrete hyper-parameter configuration in the process. 
We then showcase the potential of the presented approach in a classification task, in which the algorithm is found to perform comparably to a SOM. 
Additionally, we demonstrate that the prescribed hyper-parameter configuration is not strongly dependent on map size, or on the structure of the data - see subsections \ref{exp:scalability} and \ref{applicability}. 
Finally, we are able to analytically derive an upper bound for the computational complexity \ref{complexity} under the prescribed hyper-parameterization, which is found to equal that of synchronously trained approaches in the domain (SOM).

Future works may focus on enhancing the components of the proposed algorithm, e.g. on a more effective distributed search method. 
Another option would be to consider alternative cascading mechanisms, that can be subjected to faster external drive - which would result in higher data throughput (that is, more samples may be processed simultaneously on different units). Other works may focus on industrial applications which may benefit by increasing the maximum tractable map size.

\bibliography{DMAN}
\bibliographystyle{icml2020}


\appendix
 
 \newpage
\section{Appendix}\label{appendix}
\begin{figure*}[ht!]
	\centering
	\includegraphics[scale=0.5]{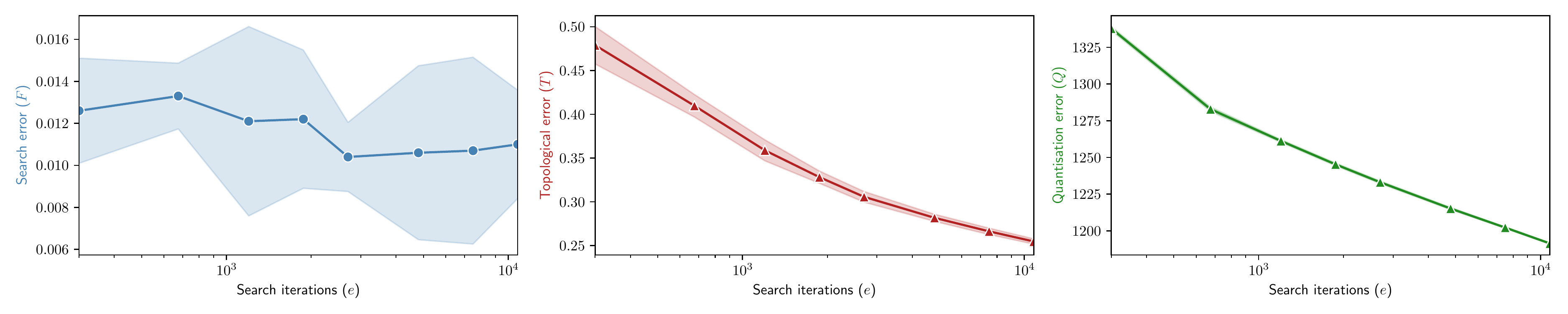}
	\caption{The efficacy of the presented heuristic search scheme in not sensitive to map size  $N$ - which allows for the map quality to improve as $N$ increases.
		The left subplot reveals that the search error remains around the same level, while the centre and right subplots reveal that topological and quantisation errors decrease along with $N$. 
		Error bands are placed at one standard deviation, and are not visually discernable for the rightmost (green) subplot. 
		$10$ maps were trained for each $N$ value, while the number of search iterations is set to $e=3N$. 
		The remaining details of the experiment setup can be found in Section 3, paragraph "Default configuration".}\label{appendix1}
\end{figure*}

We assess the scalability of the heuristic search (see Subsection 2.1) - under the proposed hyperparameterisation scheme (see Section 3, paragraph `Default configuration') - experimentally.
Specifically, we use the MNIST dataset and calculate the search error (defined in Subsection 2.1), and topological \& quantisation errors (defined in Section 3) for maps of inreasing sizes ($N \in \{ 100, 225, 400, 625, 900, 1600, 2500, 3600 \} $). 
For each $N$ value we simulate $10$ maps. 

The results of the experiment are depicted in figure \ref{appendix1}, which depicts the average search, topological, and quantisation errors - along with the respective error bands. 
The leftmost subplot demonstrates that search error remains on the same level for increasing $N$, which implies that the proposed search is scalable with respect to system size (for the given hyperparametrisation and dataset). 
Additionally, the centre and right subplots depict both quantisation and topological error decreasing along with $N$. 
These observations lead us to the conclusion that the proposed search scheme is scalable with respect to size - in the sense that the map's topological and approximation quality to improve as $N$ increases.

\end{document}